\documentclass[10pt,twocolumn,letterpaper]{article}

\usepackage{wacv}
\usepackage{times}
\usepackage{epsfig}
\usepackage{graphicx}

\usepackage{amsmath}
\usepackage{amssymb}
\usepackage{enumitem}
\usepackage{braket,booktabs}
\usepackage{pifont}
\newcommand{\cmark}{\ding{51}}
\newcommand{\xmark}{\ding{55}}
\usepackage{multirow}
\usepackage{pbox}
\usepackage[linesnumbered,ruled,vlined,boxed]{algorithm2e}
\usepackage{textcomp}
\usepackage{floatrow}
\newfloatcommand{capbtabbox}{table}[][\FBwidth]
\usepackage{subfigure}
\usepackage{array}

\def\wacvPaperID{988} 

\wacvfinalcopy 

\ifwacvfinal
\def\assignedStartPage{1}
\fi

\ifwacvfinal
\usepackage[breaklinks=true,bookmarks=false]{hyperref}
\else
\usepackage[pagebackref=true,breaklinks=true,colorlinks,bookmarks=false]{hyperref}
\fi

\ifwacvfinal
\else
\fi

\newcommand{\Te}{\emph{Teacher}}
\newcommand{\te}{\emph{teacher}}
\newcommand{\St}{\emph{Student}}

\begin{document}

\title{Effectiveness of Arbitrary Transfer Sets for Data-free Knowledge Distillation}

\author{Gaurav Kumar Nayak\thanks{denotes equal contribution.}\\ 
Indian Institute of Science\\
Bangalore, India\\
{\tt\small gauravnayak@iisc.ac.in}
\and
Konda Reddy Mopuri\footnotemark[1]\\
Indian Institute of Technology\\
Tirupati, India\\
{\tt\small kmopuri@iittp.ac.in}

\and
Anirban Chakraborty\\
Indian Institute of Science\\
Bangalore, India\\
{\tt\small anirban@iisc.ac.in}
}

\maketitle

\begin{abstract}
   Knowledge Distillation is an effective method to transfer the learning across deep neural networks. Typically, the dataset originally used for training the \textit{Teacher} model is chosen as the ``Transfer Set'' to conduct the knowledge transfer to the \textit{Student}. However, this original training data may not always be freely available due to privacy or sensitivity concerns. In such scenarios, existing approaches either iteratively compose a synthetic set representative of the original training dataset, one sample at a time or learn a generative model to compose such a transfer set. However, both these approaches involve complex optimization (GAN training or several backpropagation steps to synthesize one sample) and are often computationally expensive. In this paper, as a simple alternative, we investigate the effectiveness of ``\textit{arbitrary transfer sets}'' such as random noise, publicly available synthetic, and natural datasets, all of which are completely unrelated to the original training dataset in terms of their visual or semantic contents. Through extensive experiments on multiple benchmark datasets such as MNIST, FMNIST, CIFAR-10 and CIFAR-100, we discover and validate surprising effectiveness of using arbitrary data to conduct knowledge distillation when this dataset is ``target-class balanced''. We believe that this important observation can potentially lead to designing baselines for the data-free knowledge distillation task.
\end{abstract}

\section{Introduction}
\label{sec:intro}
Knowledge Distillation (KD)~\cite{bucilua2006model,hinton2015distilling} is a contemporary technique for transferring learning across neural network models. Typically, knowledge from one or more complex and deep models (called \Te s) is distilled into a relatively lightweight model (called \St{}). The core idea of Knowledge Distillation, as discussed in the seminal paper by Hinton \etal~\cite{hinton2015distilling}, is to transfer the (input to output) learned mapping function from \Te{} to \St{} via sharing the ``dark knowledge'' extracted by the \Te{} on the training images. This typically is achieved via matching the soft targets (or soft labels, i.e., output of softmax layer) predicted by the \St{} to that of the \Te{} for the same inputs. This is the distillation mechanism that enables transfer of the better generalization capability (i.e., the ``knowledge'') of the \Te{} to the \St{}. Thus, Knowledge Distillation has established itself as a very useful and practical tool because of its simplicity and potential.
\begin{figure*}[h!]
\hfill
\subfigure[]{\includegraphics[width=4.50cm]{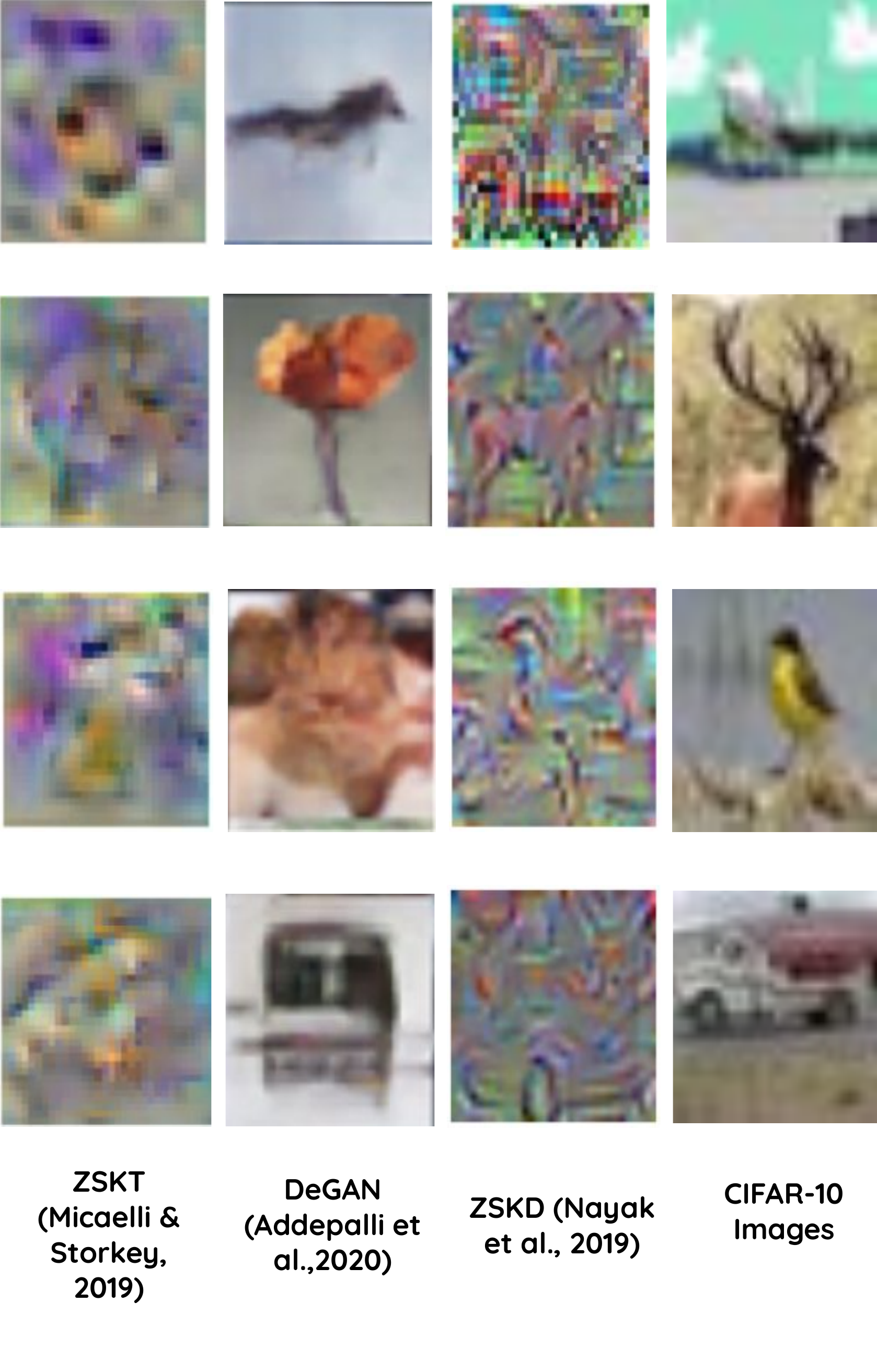}
\label{fig:pseudo-samples}}
\hfill
\subfigure[]{\includegraphics[width=0.65\columnwidth, height=7.6cm]{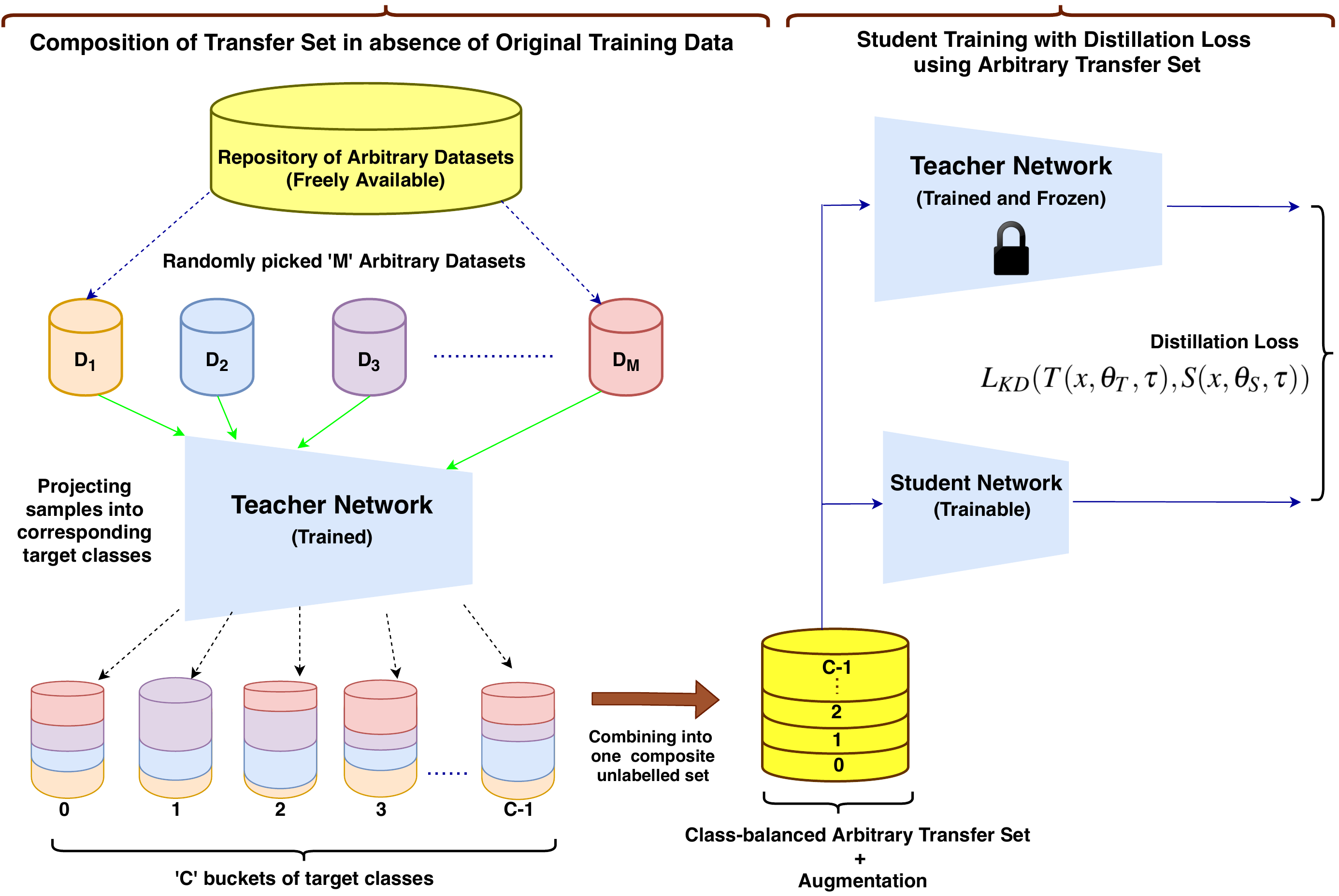}
\label{fig:overview}}
\hfill
\caption{(a) Example of pseudo samples generated by existing data-free KD approaches for the CIFAR-10 dataset compared against actual target data samples (rightmost column), (b) Our proposed KD Baseline: An approach depicting simple and effective way of performing KD in absence of the original training data by utilizing arbitrary data samples to construct the transfer set.}
\end{figure*}


The samples used for performing distillation constitute the ``Transfer set", which is typically required to be constructed using the data sampled from the target distribution. Therefore, the most commonly used transfer set is the original training dataset on which the Teacher model was trained. However, this requirement has been identified as a limitation (e.g.~\cite{zskd-icml-2019,zskt-neurips-2019}) since it is common now-a-days that many popular pre-trained models are released without providing access to the training data (e.g. Facebook's Deepface model trained on $4$M confidential face images). This is due to one or more practical constraints such as (i) privacy (e.g. models trained on patients' data from hospitals), (ii) property (proprietary data of companies that invest on collection and annotation), and (iii) transience (observations from the training of a reinforcement learning environment do not exist). 

To handle this ``data-free'' (or zero-shot) distillation scenario, most of the approaches broadly follow either of the two ways: (i) compose a synthetic transfer set by directly utilizing the trained \Te{} model that acts as a proxy to the target data (e.g.~\cite{zskd-icml-2019,dfkd-nips-lld-17}), or (ii) capture the target data distribution using generative models (e.g.~\cite{zskt-neurips-2019,dafl-iccv-2019,degan-aaai-2020}). Both these approaches suffer from heavy computational overhead: iteratively crafting synthetic samples via several steps of backpropagation through the \Te{} or learning a complex GAN like generator framework that involves complicated optimization. Some of these approaches (e.g.~\cite{dfkd-nips-lld-17}) additionally need to store meta-data about the original training dataset (e.g. feature statistics of the \Te{} model) for generating the synthetic transfer set. Further, in case of image data, the generated samples are often observed to be visually quite dissimilar (Fig.~\ref{fig:pseudo-samples}) to the training data samples. That means, they do not lie close to the training samples in the data manifold. At the same time, it is unclear how or why these samples, despite seemingly being ``out-of-distribution'' and ``far-from-real'', enable effective transfer between the models, as evidenced by the reported results.

These observations motivate us to investigate the effectiveness of any \textit{arbitrary transfer set} towards the task of knowledge distillation, despite it being unrelated to the original training data. If proven effective, such datasets can in fact be used to design important and often strong baselines for the KD tasks, while saving us the large overhead of composing synthetic transfer sets, as incurred by the existing data-free distillation approaches. This is especially true for text/image domains, where it is easy to collect large volume of unlabeled arbitrary data from ubiquitous publicly available sources. More importantly, this investigation can uncover important insights into the mechanism of the distillation process.

Therefore, in this work, we consider a wide range of unlabelled stimuli from different content worlds in the context of distillation. Specifically, we consider (i) random noise inputs, (ii) arbitrary synthetic datasets, and (iii) arbitrary natural datasets towards composing the transfer set. However, it is observed (refer to Sec. \ref{subsec:kd-absence}) that the deep neural networks often partition the arbitrary input domain into disproportionate classification regions. In other words, arbitrary data samples may not be projected uniformly into the learned classification regions of the \Te{}. This imbalance in the classification regions results the \St{} to overfit the classification boundaries during the distillation. 
In other words, it can not preserve the class decision boundaries learnt from the original training data, thereby seriously affecting the \St{}'s generalization performance. 
These observations lead to the hypothesis that an ideal transfer set should equally represent all the classification regions of the \Te{} model which can minimize the distortion in decision boundary and hence would help in achieving effective knowledge transfer. In other words, the arbitrary transfer set needs to be ``target-class balanced'' in order to successfully impart \Te{}'s learning to the \St{}. 

In summary, the contributions of this work are, as follows:
\begin{itemize}
\itemsep0em
\item For the first time in the literature, we show that arbitrary transfer sets, unrelated to the target data set, can be effectively utilized 
for the task of knowledge distillation in the ``data-free'' scenario. 
\item To maximize the efficacy of distillation using such transfer sets, we present a simple yet effective approach of making them ``class-balanced''. 
\item We empirically demonstrate the effectiveness of the proposed approach on multiple benchmark datasets such as MNIST, FMNIST, CIFAR-10 and CIFAR-100, as we achieve performance comparable to state-of-the-art data-free distillation approaches. 
\end{itemize}
\begin{figure*}[htp]
\centering
\centerline{\includegraphics[width=\textwidth]{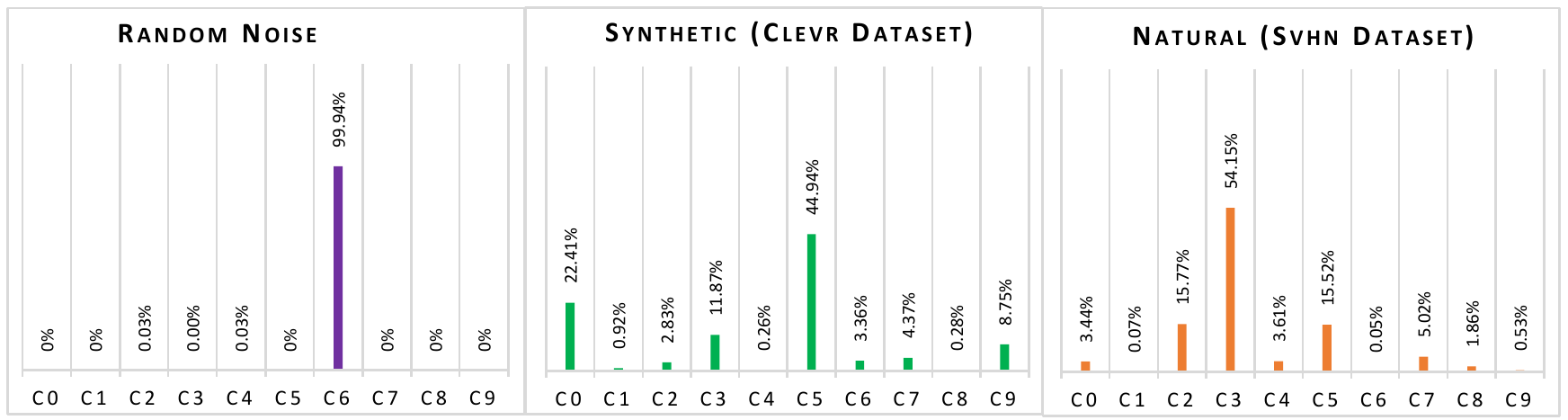}}
\caption{Percentage of the total number of arbitrary samples distributed over the set of the target classes (of CIFAR-10) by the trained \Te{} model for different arbitrary datasets.}
\label{fig:arbitrary-label-distribution}
\end{figure*}

\section{Related Works}
\label{sec:related}
Our work is broadly related to the data-free Knowledge Distillation. Early works (e.g.~\cite{bucilua2006model,hinton2015distilling}) use the entire training data as the transfer set. Buciluǎ~\textit{et al.}~\cite{bucilua2006model} suggest to meaningfully augment the training data for effectively transferring the knowledge of an ensemble onto a smaller model. Recently, there have been multiple approaches to perform knowledge transfer in the absence of training data. They can be broadly categorized into two branches: (i) methods that extract samples related to the training data from the \Te{} model, (ii) methods that attempt to learn the training distribution (e.g., using GAN-like generative models). 

General idea of the first category is to iteratively modify a randomly initialised stimulus (input) via back propagation in order to  maximize the \Te 's class confidence. Nayak~\textit{et al.}~\cite{zskd-icml-2019} compose the synthetic transfer set, which is carefully crafted by modelling the soft-label space. Lopes~\textit{et al.}~\cite{dfkd-nips-lld-17} and Bharadwaj~\textit{et al.}~\cite{dreamdistillation-iclrw-2019} save information in the form of \Te{}'s feature statistics in order to acquire transfer set that is closer to the training data. They further add diversity to the set via adding small noise to the saved statistics and generating the samples. These methods are computationally expensive requiring thousands of back propagation iterations per sample. Further, some of them require to store the feature statistics of the \Te{} model which may not be available. 

Another direction of research for handling the absence of training data is to train a generative model that can seed the proxy samples. \cite{zskt-neurips-2019,dafl-iccv-2019,degan-aaai-2020} show that properly optimized generative models can generate samples to be strongly classified by the \Te{} models. After learning such GAN-like models, generated samples can be used as a transfer set for performing the Knowledge Distillation. Despite generating \textit{far from real} and \textit{out-of-distribution} samples (Figure~\ref{fig:pseudo-samples}) these methods are observed to successfully do the knowledge transfer. \cite{degan-aaai-2020} attempts to use a proxy natural dataset for influencing the generations to be similar to proxy data distribution while bringing its features close to original training data distribution. These methods involve training generative models that require careful balancing of multiple terms in the objective. In this work, we take a different direction to study the effectiveness of cheaply available, unlabelled arbitrary data as transfer set. We put forward an intuitive strategy to compose effective transfer sets that can yield potential baselines and empirically demonstrate its efficacy.
\section{Class-balanced Arbitrary Transfer Sets}
\label{sec:proposed}
We first briefly review the principles of Knowledge Distillation, and subsequently introduce our framework to compose an effective transfer set from unlabelled arbitrary data sources.
\subsection{Knowledge Distillation (KD)}
\label{subsec:kd}
Knowledge Distillation typically uses the original training data as the transfer set on which the \Te{} model is trained. Let us denote the \Te{} as $T$, \St{} as $S$, their parameters as $\theta_T$ and $\theta_S$ respectively, and the transfer set as $\mathcal{D}$ that consists of the input-target tuples denoted by $(x,y)$. Note that typically \St 's capacity would be smaller compared to that of \Te{}, i.e., $|\theta_S| \ll |\theta_T|$. The objective of Knowledge Distillation is to train the \St{} in order to match the soft labels produced by the \Te{} along with learning to predict the correct hard labels on the training set. This objective can be realized via minimizing
\begin{equation}
    L=\sum_{ (x,y) \in \mathcal{D} }L_{KD}(S(x,\theta_S,\tau),T(x,\theta_T,\tau))  +  \lambda \cdot L_{CE}(\hat{y}, y)
    \label{eqn:kd}
\end{equation}
where, $L_{KD}$ is distillation (e.g. $l_2$ or cross-entropy) loss computed between the soft labels of $T$ and $S$, $L_{CE}$ is cross-entropy loss comparing the ground truth $(y)$ with the prediction $(\hat{y})$ by $S$, $\tau$ is the temperature used in distillation, and $\lambda$ is a hyper-parameter balancing the loss terms.
\subsection{KD with Arbitrary Transfer Set}
\label{subsec:kd-absence}
In absence of the original training data (refer to sec.~\ref{sec:intro} for such scenarios) multiple approaches (e.g.~\cite{zskt-neurips-2019,zskd-icml-2019,dreamdistillation-iclrw-2019,degan-aaai-2020}) compose synthetic transfer set and achieve effective distillation. However, it is clear (Figure~\ref{fig:pseudo-samples}) that these samples are visually very different from the training samples and hence may not actually lie on the data manifold.

Motivated from these observations, we consider investigating the effectiveness of transfer sets composed of arbitrary samples towards conducting KD. That is, if the transfer sets are composed via random picking of samples from a limited supply of publicly available datasets, as opposed to careful crafting or selection. For instance, in case of the object recognition models trained on CIFAR-10~\cite{krizhevsky2009learning} target dataset, Clevr~\cite{johnson2017clevr} and SVHN~\cite{netzer2011reading} are a pair of candidate arbitrary datasets. Note that the later two are unrelated to the target dataset, i.e., they do not share category labels or similar visual/semantic information with the target dataset.

When we compose a transfer set, generally we attempt to ensure that there are samples from all the classification regions of the \Te{}. However, it is unlikely that an arbitrarily composed transfer set will have samples from all the classification regions representing the data distribution. That means, the distribution of labels predicted by the \Te{} can be extremely unbalanced. As a consequence, the decision boundaries learnt by the \St{} model using arbitrary data as a transfer set will be distorted with respect to that learnt using the original training samples (i.e. boundaries learnt by the \Te{}). 
For instance, Figure~\ref{fig:arbitrary-label-distribution} shows the distribution of labels predicted by AlexNet \Te{} trained on CIFAR-10 for three different arbitrary datasets: Random noise, Clevr and SVHN. It can be noticed that these distributions are far from uniform. This might be attributed to the disproportionate classification regions labelled by pretrained deep models on arbitrary transfer sets. Clearly, it is unlikely that arbitrary data would be able to preserve the same class boundaries learnt using the original training data. Hence, the randomly composed transfer sets are not ideal for performing the distillation (section~\ref{sec:experiments}).

In light of such important observations, we propose a simple but effective strategy to ensure representation of all the classification regions in the transfer set, which helps to mitigate the distortion in the decision boundaries. 
While composing an arbitrary transfer set, we enforce it to have a label distribution closer to uniform over the set of target labels by design. Note that the predicted labels would still be completely unrelated to the visual patterns, i.e., one may not expect any semantic/visual similarity between a datapoint in the arbitrary transfer set and the original training set even for the same predicted label. 
Despite being unrelated samples, we aim to have these spanned uniformly across all the classification regions, thereby forming the `Target class-balanced' arbitrary transfer set.
\begin{algorithm}[t]
{
\caption{KD with class balanced arbitrary transfer set}
\label{algo}

\SetAlgoLined
\SetKwInOut{Input}{Input}  
\Input{\Te{} \textit{T}, \St{} \textit{S}, arbitrary datasets : $\{D_1, D_2, \ldots D_M\}$, intended maximum size of the transfer set $N$}
\SetKwInOut{Output}{Output}  
\Output{
Trained Student model weights $\theta_S$, \\
$\bar{D}$: \emph{Target-class balanced} Transfer set}

Obtain $C$: Number of categories from the output-space dimension of \emph{T}, and Initialize $\bar{D} = \phi$

Initialize sample counts of each target class in $\bar{D}$:  $c_i = 0$, $ \forall \:\:\: i \:\:\: \in \:\:\: \{0,1,\ldots C-1 \}$

 \For{i=1:M}{
    
        \While{$\exists \:\: j $ \emph{such that} $c_j < \lfloor N/C \rfloor$ \emph{and} $D_i \neq \phi$}{
        
        sample $x_k \in D_i$ 
        
        $D_i \leftarrow D_i \setminus \{x_k\}$

        $l \leftarrow$ class-label predicted by $T$ on $x_k$
        
        \If{$c_l < \lfloor N/C \rfloor $}{
        
        $\bar{D} \leftarrow \bar{D} \cup \{x_k\}$ 
        
        $c_l \leftarrow c_l + 1$ 
        
        }
        
        }
        
        \lIf{$c_j = \lfloor N/C \rfloor \:\: \forall \:\: j \:\: \in \{0,1,\ldots C-1\}$}{break}
        
        
        
    }
  Perform the Distillation via optimizing for $\theta_S^*=\underset{\theta_S}{\mathrm{\:argmin\:}} \sum_{x \in \bar{D}} L_{KD}(T(x,\theta_T,\tau), S(x,\theta_S,\tau))$ 
}
\end{algorithm}
\begin{figure*}[htp]
\centering
\centerline{\includegraphics[width=\textwidth]{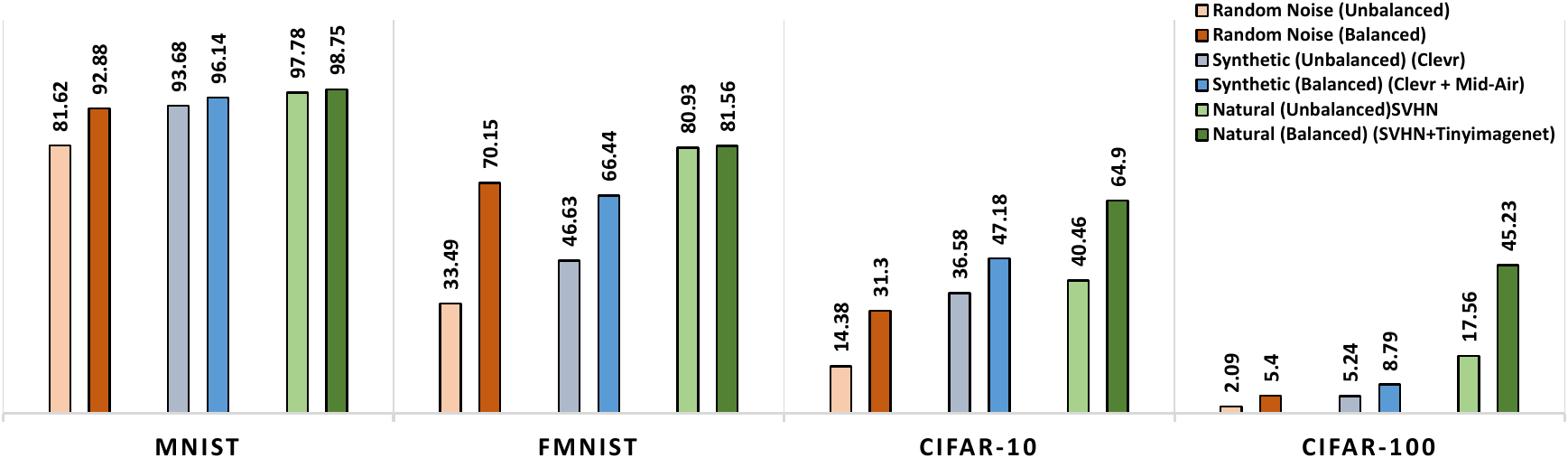}}
\caption{Comparison of the distillation performance using unbalanced and balanced arbitrary transfer sets. Balanced set outperforms its unbalanced counterpart across all the three different varieties of arbitrary datasets: noise, synthetic and unrelated natural data.}
\label{fig:balance v/s unbalance}
\end{figure*}
Algorithm~\ref{algo} (also Figure~\ref{fig:overview}) shows the steps for composing such a transfer set from a repository of freely available unlabelled datasets using our hypothesis. Note that it may not be possible to get exactly uniform predicted label distribution with the finite supply of arbitrary samples. However, the goal here is to carefully avoid the aforementioned extreme imbalance that creeps in with a random composition. Unlike existing data-free KD approaches such as~\cite{zskd-icml-2019, dafl-iccv-2019, degan-aaai-2020}, we do not generate any synthetic samples and use the existing unlabelled datasets in their original forms. While composing the arbitrary transfer set, we only require a single forward pass for each sample through the \Te{} model and there are no backpropagations involved which makes our proposed algorithm much less compute-intensive, especially when compared to existing methods \cite{zskd-icml-2019, dafl-iccv-2019, degan-aaai-2020}. After composing such a transfer set $(\bar{D})$, we perform distillation via optimizing the \St{} model $(\theta_S)$
\begin{equation}
\theta_S^*=\underset{\theta_S}{\mathrm{\:argmin\:}} \sum_{x \in \bar{D}} L_{KD}(T(x,\theta_T,\tau), S(x,\theta_S,\tau))     
\label{eqn:kd-arbitrary}
\end{equation}
where $L_{KD}$ is the distillation objective (cross entropy loss is used in our experiments), $\tau$ is the temperature used in softmax layers of $T$ and $S$. Note that, unlike eq.~\ref{eqn:kd}, eq.~\ref{eqn:kd-arbitrary} does not contain the classification loss ($L_{CE}$) because the transfer set is arbitrary, i.e., unrelated to the target classes and hence forcing hard labels on these samples is counter-intuitive.
\section{Experiments}
\label{sec:experiments}
In this section, we empirically demonstrate the importance of target-balanced arbitrary transfer sets as a strong baseline for performing distillation. Before we present the experimental results, we describe the CNN classifiers and the datasets we used in our experiments. 

\textbf{MNIST/FMNIST and LeNet-5}: The MNIST \cite{lecun1998gradient} dataset contains images of handwritten digits. FMNIST \cite{xiao2017fashion} has images of several fashion items. Both the datasets contain $60000$ training images, $10000$ test gray scale images. Lenet-$5$ is taken as \Te{} and Lenet-$5$-Half as \St{} (identical setting to \cite{dfkd-nips-lld-17, zskd-icml-2019}).
 
\textbf{CIFAR-10 and AlexNet/ResNet}: CIFAR-10 \cite{krizhevsky2009learning} dataset has colour images of size $32 \times 32$. AlexNet~\cite{krizhevsky2012imagenet} is taken as \Te{} and AlexNet-Half as \St{} to have a fair comparison with \cite{zskd-icml-2019, degan-aaai-2020}. Similar to \cite{dafl-iccv-2019}, we also perform experiments with Resnet-34 as \Te{} model and Resnet-18 as \St{} which are bigger networks in comparison to AlexNet and AlexNet-Half.
 
\textbf{CIFAR-100 and Inception-v3}: In order to demonstrate the validity of our hypothesis even on large scale datasets, we also experiment on CIFAR-100~\cite{krizhevsky2009learning} which is similar to CIFAR-10 but contains $100$ classes instead of $10$. Train data has $500$ images per class while test data contains $100$ images per class. Similar to \cite{degan-aaai-2020}, we take Inception-V3 \cite{szegedy2016rethinking} as \Te{} and ResNet-18 \cite{he2016deep} as \St{}.

We use the following datasets as \emph{Transfer} sets in the absence of original training data:
 
\textbf{Random noise}: Uniform random noise in $[0,1]$ for each pixel in an image is used to construct transfer set for MNIST, FMNIST and CIFAR-10. For CIFAR-100, we found that creating balanced set using Gaussian noise $(\mu=0.5, \sigma=0.1)$ (clipped to $[0,1]$) is faster.

\textbf{Synthetic datasets}: We use publicly available synthetic dataset `Clevr'~\cite{johnson2017clevr} as unbalanced transfer set. This dataset contains $70000$ images. In order to improve the class-balance of Clevr, we add another synthetic dataset, called Mid-Air~\cite{Fonder2019MidAir} on top of it. This dataset contains images under different climate conditions. We resize each image to $32 \times 32$.

\textbf{Natural datasets}: We utilize SVHN~\cite{netzer2011reading} dataset as unbalanced arbitrary natural transfer set. It has $73257$ images of street view house numbers. We use the cropped version of the dataset where each image is resized to $32 \times 32$. To achieve class balance while keeping the ``naturalness'' of images in the transfer set, we add samples from TinyImageNet~\cite{le2015tiny} on top of it.
\subsection{Arbitrary Transfer Sets for Distillation}
\label{subsec:arbitrary-transfer-sets}
In all our experiments, size of the transfer set is kept approximately equal to that of corresponding \Te{}'s training set. This enables a fair comparison of the distillation performance between arbitrary and training datasets. The exact size of the transfer sets used in the experiments along with exact number of arbitrary samples labelled in each of the target classes is provided in the supplementary material. We present and analyse the experimental results separately for each of the different stimuli.

\begin{table*}[htp]
\centering
{\small
\begin{tabular}{|l|c|c|c|c|c|c|c|c|c|} \hline
          \multirow{2}{*} {\textbf{Transfer Set}} &  \multirow{2}{*}{\textbf{Balanced}} & \multicolumn{2}{c|}{\textbf{MNIST}} & \multicolumn{2}{c|}{\textbf{FMNIST}} & \multicolumn{2}{c|}{\textbf{CIFAR-10}} & \multicolumn{2}{c|}{\textbf{CIFAR-100}} \\  \cline{3-10}
          &  & w/o Aug      & w/ Aug     & w/o Aug & w/ Aug     & w/o Aug       & w/ Aug       & w/o Aug        & w/ Aug       \\ \hline
Random Noise &   \xmark  & 81.62           &   89.01           &    33.49           &   38.37           &     14.38           &   47.50          &  2.09            &  3.30 \\ 
Random Noise & \cmark   &   92.88         &   95.76           &     70.15          &    74.33         &    31.30            &      67.40      &      5.40        &  18.20\\ \hline \hline
Clevr & \xmark &     93.68        &    97.11        &    46.63           &    72.68          &             36.58   &    72.35        &       5.24       &  17.45 \\
Clevr+Mid-Air & \cmark &     96.14         &     98.53       &     66.44          & 83.38             &              47.18  &   75.76            &    8.79        & 22.28\\ \hline \hline
SVHN & \xmark  &    97.78          &    98.81        &     80.93          &    83.85          &    40.46            &    72.33          &    17.56          &  40.59\\ 
SVHN + Tiny &  \cmark  &      98.75        &     98.96       &       81.56        &    84.75          &   64.90             &     79.19        &       45.23       &  67.18 \\

\hline
\end{tabular}
}
\caption{Effect of Augmentation: Distillation performance using unlabelled arbitrary transfer sets on multiple datasets with and without augmentation (Tiny stands for `TinyImageNet').}
\label{tab:kd-arbitrary-data}
\end{table*}
\begin{table*}[htp]
    \centering
    {\small
    \begin{tabular}[t]{|l|c|c|c|c|}
        
        \hline
        \textbf{Algorithm} & \textbf{MNIST} & \textbf{FMNIST} & \textbf{CIFAR-10} & \textbf{CIFAR-100}     \\ \hline \hline
        \Te{} & 99.34 & 90.84 & 83.03 & 79.05 \\ \hline
        Student-KD~\cite{hinton2015distilling} & 99.25 & 89.66 & 81.78 & 69.65 \\ \hline \hline
        ZSKD~\cite{zskd-icml-2019} &   98.77    &    79.62    &    69.56      &     $-$      \\ \hline
        DeGAN~\cite{degan-aaai-2020} &   $-$    &   83.79     &  \textbf{80.55}      &    65.25       \\ \hline

        \pbox[c][0.95cm][c]{4.0cm} {Ours\\(SVHN + TinyImageNet) }   &  \textbf{98.96}     &    \textbf{84.75}    &    79.19      &      \textbf{67.18}     \\ \hline
        
    \end{tabular}\hfill%
    \begin{tabular}[t]{|l|c|}
    
        \hline
         \textbf{Algorithm} & \textbf{CIFAR-10} \\\hline \hline
        \Te{} (ResNet 34) & 95.58 \\ \hline
        
        Student-KD (ResNet 18)~\cite{hinton2015distilling} & 94.34 \\ \hline \hline
        DAFL~\cite{dafl-iccv-2019} & 92.22 \\ \hline
        \pbox[c][.95 cm][c]{4cm}{Ours\\(TinyImageNet + SVHN)}  & \textbf{92.92} \\  \hline
        
    \end{tabular}%
    }
\caption{Comparison with SOTA : Performance of proposed method in comparison with ZSKD \cite{zskd-icml-2019} and DeGAN \cite{degan-aaai-2020} (table on the left), \& DAFL \cite{dafl-iccv-2019} (table on the right)}
\label{tab:comparison-with-sota}
\end{table*}

\textbf{Random noise stimuli}:
We consider the following two scenarios: (i) Unbalanced: randomly sampled noise samples, (ii) Balanced: noise set on which \Te{} predicts all the target labels almost uniformly. In Fig.~\ref{fig:balance v/s unbalance}, the first two bar graphs for each dataset show the distillation performance of the random noise transfer set. In the case of MNIST, we observe that even an arbitrary random noise gives a decent performance of $81.62\%$. By class-balancing the random noise, we can increase the accuracy to $92.88\%$. In the case of FMNIST and CIFAR-10 we get a significant improvement of close to $36\%$ and $17\%$ respectively on target-class balancing while $3.3\%$ gain in case of CIFAR-100.

\textbf{Synthetic stimuli}: For the unbalanced case we consider the Clevr dataset to perform the distillation without looking at their predicted labels by the \Te{}. For the class-balanced case, we add samples from another synthetic dataset, Mid-Air, towards obtaining approximately equal number of samples in each of the target classes, as described in Algorithm~\ref{algo}. 
The third and fourth bar graphs (for each dataset) in Figure~\ref{fig:balance v/s unbalance} show the distillation performance of the synthetic stimuli. We get a decent improvement of $2.46\%$ and $3.55\%$ in case of MNIST and CIFAR-100 while significant gain of $19.81\%$ and $10.60\%$ for FMNIST and CIFAR-10 respectively by using a class-balanced Synthetic dataset. 

\textbf{Natural image data stimuli}: We consider SVHN as the arbitrary transfer set. For achieving class-balance, we use samples from TinyImageNet which are added on top of SVHN as described in Algorithm~\ref{algo}. The last two bar graphs for each dataset in Figure~\ref{fig:balance v/s unbalance} show the results with natural data as transfer set. We get a small gain of $0.97\%$, $0.63\%$ for MNIST and FMNIST due to the (relatively) low imbalance while a significant gain of $24.44\%$ and $27.67\%$  for CIFAR-10 and CIFAR-100 respectively due to high imbalance in target classes, where class balancing has a profound effect.
\subsection{Augmentation Helps the Underrepresented Classes}
\label{subsec:effect-of-augmentation}
Note that in all the three scenarios, practically it is very difficult to achieve perfect class-balance (identical number of samples in each target class) with limited supply of arbitrary data. We noticed that more frequently the classification regions learned by the deep models are heavily out of proportion and that makes it difficult to have arbitrary samples representing all the target classes equally. However, the objective of achieving better distillation performance is to reduce the extreme class imbalance and compose a transfer set that represents all the classification regions. Therefore, after achieving some level of balance (as in Algorithm~\ref{algo}), we further improve the representation from the under populated classes via performing augmentations during the distillation process. 
Our augmentation includes scaling, rotation, flipping, adding random noise (Gaussian, salt and pepper), and multiple combinations of them. Although augmentations add diversity to all the target classes, underrepresented classes get more benefited than the relatively better populated ones resulting in further less distortion of the class decision boundaries and hence, improved distillation performance. Table~\ref{tab:kd-arbitrary-data} shows the results with and without augmentation across all the transfer scenarios over multiple datasets. It is evident that augmentation consistently results in better distillation performance via boosting the underrepresented classes in the transfer set.

\subsection{Comparison With the State-of-the-art}
\label{subsec:comparison-with-sota}
\begin{figure}[htp]
\centering
\centerline{\includegraphics[width=0.87\textwidth, height=0.47\textwidth]{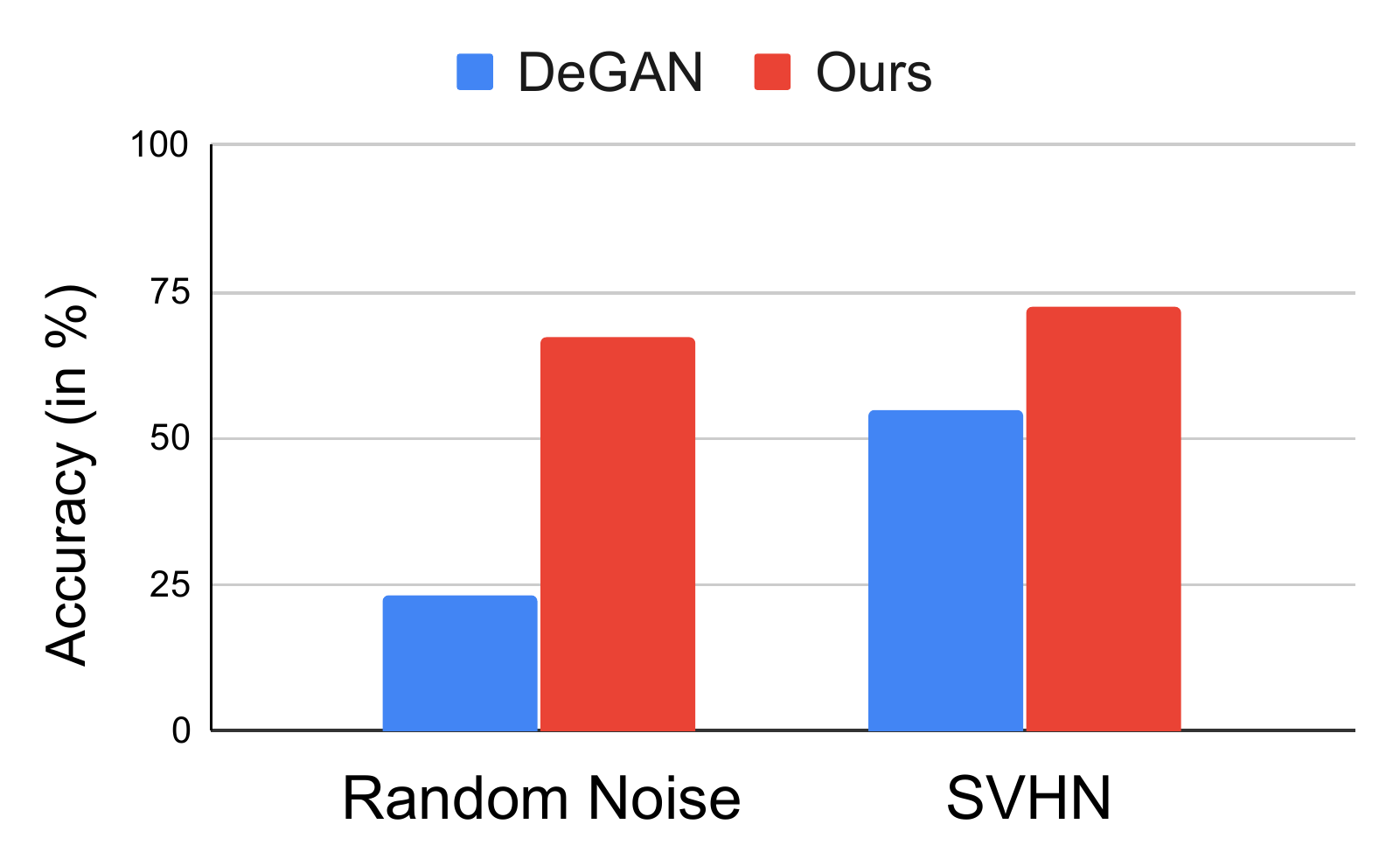}}
\caption{Comparison of our proposed approach with DeGAN~\cite{degan-aaai-2020} when unrelated transfer sets are used to distill the knowledge from \Te{} model trained with CIFAR-10.}
\label{fig:sota}
\end{figure}

In this subsection, we compare the proposed baseline against the state-of-the-art data-free knowledge distillation approaches. Table~\ref{tab:comparison-with-sota} (on left) presents the comparison of distillation performance of the proposed approach against ZSKD~\cite{zskd-icml-2019} and DeGAN~\cite{degan-aaai-2020} on multiple datasets. In order to have a fair comparison, we experimented with the same models used in~\cite{zskd-icml-2019} and~\cite{degan-aaai-2020}. The proposed baseline clearly performs better on MNIST, FMNIST and CIFAR-100. In case of CIFAR-10, it is close to DeGAN's~\cite{degan-aaai-2020} performance. But, DeGAN achieves better performance on CIFAR-10 when samples from the classes of CIFAR-100 are used. These CIFAR-100 samples are substantially more similar to CIFAR-10 than our arbitrary dataset (SVHN$+$TinyImageNet), thereby resulting in this improved distillation performance. If an unrelated dataset is used for distillation from the \Te{} network on CIFAR-10, proposed baseline outperforms the DeGAN significantly (as shown in Figure~\ref{fig:sota}). 
We have also compared our proposed baseline with DAFL~\cite{dafl-iccv-2019}. Again, to have a fair comparison, we have used the ResNet-34 \Te{} and ResNet-18 \St{} used in~\cite{dafl-iccv-2019}. The performance comparison is shown in Table~\ref{tab:comparison-with-sota} (on right). Our baseline on arbitrary/unrelated transfer set performs slightly better. Further, unlike DAFL, it does not require any complicated GAN training.
\subsection{Strict Arbitrary Transfer Sets : Explicit Removal of Overlapping Classes}
\label{subsec:remove-non-overlapping}
In this subsection, we explicitly make sure that the arbitrary transfer sets do not have any overlap with the target categories and is semantically very dissimilar with the target data samples. We investigate the true potential of such strict arbitrary transfer sets when they are \textit{target-class balanced} and used for knowledge distillation in complete absence of original training data.

We chose TinyImageNet as arbitrary data in section \ref{subsec:arbitrary-transfer-sets} due to its relatively larger size, which can help to make the transfer set better balanced. Also, this dataset is a widely used publicly available dataset. However, it may contain a few overlapping classes with CIFAR. Therefore, we further perform experiments to show that the distillation performance is close to that using TinyImageNet even if we consider arbitrary datasets that do not share any categories. 

\begin{table}[htp]
\centering
{\small
\begin{tabular}{|l|c|}
\hline
\textbf{Transfer Set}                                                                               
& \textbf{Distillation Accuracy}  
     \\ 
     \hline \hline
\begin{tabular}[c]{@{}l@{}}SVHN (Digits) + \\ CelebA (celebrity faces)\end{tabular}        
& 75.87 \%                    \\ \hline
\begin{tabular}[c]{@{}l@{}}SVHN + CelebA +\\ Fruits360 (fruits \& vegetables)\end{tabular} 
& 76.33 \%                    \\ \hline
\begin{tabular}[c]{@{}l@{}}SVHN + \\ CIFAR-100 (non-overlapping)\end{tabular}               
& 79.13 \%                    \\ \hline
SVHN + TinyImageNet                                                                        
& 79.19 \%                    \\ \hline
\end{tabular}}
\caption{Distillation performance when completely unrelated and non-overlapping arbitrary transfer sets are used to distill knowledge from \Te{} model trained on CIFAR-10.}
\label{tab:non-overlapping}
\end{table}

SVHN contains images of digits which are dissimilar to CIFAR-10 and are not balanced across target classes of trained \Te{} model on CIFAR-10. Several different unrelated datasets are added on top of SVHN to improve the target-class balance using our proposed Algorithm \ref{algo}. Instead of using TinyImageNet which may have samples related to target categories, we add CelebA \cite{liu2015faceattributes} dataset on top of it. CelebA has images of celebrity faces and does not have any overlap with CIFAR-10 classes. Please note that in all our experiments we make sure that the number of samples in arbitrary transfer set do not exceed the per class sample count of original training data in order to have a fair comparison. The results are reported in Table \ref{tab:non-overlapping} where completely unrelated arbitrary transfer sets along with augmentations are used for distillation. SVHN mixed with CelebA when used as a transfer set, gives a decent accuracy of 75.87 \%. We further observe gain in the distillation performance when the target-class balance is improved by mixing another unrelated dataset i.e. Fruits360 \cite{murecsan2018fruit} that contains images of fruits and vegetables. Please note that it is still not perfectly target-class balanced and one can further improve the accuracy by adding more non-overlapping datasets and ensuring equal amount of arbitrary samples in each of the target classes labelled by the pretrained \Te{}.

\begin{table*}[htp]
\centering
{\small
\begin{tabular}{|>{\centering} p{28mm}|>{\centering} p{25mm}|>{\centering}p{25mm}|>{\centering}p{25mm}|p{25mm}|}
\hline
\multirow{2}{*}{\begin{tabular}[c]{@{}c@{}}Size of  \\ Transfer Set (SVHN)\end{tabular}} &
  \multicolumn{2}{c|}{\textbf{Binary-MNIST} (Teacher Accuracy: 99.44\%)} &
  \multicolumn{2}{c|}{\textbf{Binary-FMNIST} (Teacher Accuracy: 93.38\%)} \\ \cline{2-5} 
      & Unbalanced     & Balanced & Unbalanced & Balanced \\ \hline \hline
16000 &   $88.89 \pm 2.05$  & $\textbf{89.70} \pm 1.86$         &   $75.04 \pm 1.83$          &    $\textbf{77.68}  \pm 1.15$       \\ \hline
8000 &   $72.60 \pm 2.41$  & $\textbf{82.69} \pm 1.73 $         &   $76.28 \pm 1.30$          &    $\textbf{78.10} \pm 0.66  $      \\ \hline
4000 &   $84.66 \pm 2.27$  & $\textbf{91.38} \pm 1.81$         &   $77.58  \pm 0.67$         &    $\textbf{79.94} \pm 0.89 $      \\ \hline
2000 &   $\textbf{83.84} \pm 3.06 $ & $82.93  \pm 3.00$         &   $68.35  \pm 2.10$         &    $\textbf{73.37} \pm 0.54  $       \\ \hline
1000 &   $81.48 \pm 1.32$ & $\textbf{82.38} \pm 1.35  $        &   $72.86   \pm 0.89$        &    $\textbf{75.21}  \pm 0.47 $       \\ \hline
500 &   $\textbf{83.67} \pm 0.25 $ & $83.12 \pm 1.23$         &   $72.95  \pm 0.70$         &    $\textbf{74.67}  \pm 0.31 $       \\ \hline
\end{tabular}}
\caption{Comparison of distillation performance (i.e., Student accuracy in \%) with unbalanced and balanced arbitrary transfer set (SVHN), when the Teacher network is trained on unbalanced binary MNIST and FMNIST training samples.} 
\label{tab:unbalanced}
\end{table*}
DeGAN~\cite{degan-aaai-2020} avoids the overlapping classes of CIFAR-100 with CIFAR-10 and reports results using the non overlapping 90 classes of CIFAR-100. For a strictly fair comparison with DeGAN, we also take the same non-overlapping CIFAR-100 samples to balance the transfer set synthesized using SVHN samples. From Table \ref{tab:non-overlapping}, we observe similar performance (SVHN+TinyImageNet) even by using the non-overlapping CIFAR-100 as the arbitrary dataset. However, since CIFAR-100 samples are added on top of SVHN, we effectively utilize only 18818 samples from CIFAR-100 as opposed to DeGAN which uses all the 45000 samples.

We, thus, empirically observe that it is possible to achieve decent distillation performance even with strictly arbitrary transfer sets (completely unrelated to and non-overlapping with original training data), when these transfer sets are target class-balanced. However, one can further improve the distillation performance by carefully selecting the arbitrary data sources by leveraging the domain knowledge and the knowledge of the task at hand while utilizing our proposed strategy (see Algo~\ref{algo}) to compose the transfer sets. 
\subsection{Unbalanced Target Dataset: Generality of the Proposed Strategy}
\label{subsec:unbalanced}
Until now we have experimented with target datasets that are class-balanced. That is, our \Te{} models are trained on equal number of samples from each class of the target dataset. In this subsection, we demonstrate the effectiveness of the balanced arbitrary transfer sets towards KD even when \Te{} is trained on unbalanced target (training) datasets. For this purpose, we have created binary classification tasks out of MNIST and FMNIST datasets separately, referred to as \textit{Binary-MNIST} and \textit{Binary-FMNIST} repectively. We merged samples from three of the ten classes (labels $0,1,$ and $2$) into one set, referred to as the `minority' class and the rest seven into another set, referred to as the `majority' class. Therefore the resulting binary classification datasets will have a $3:7$ class imbalance. We train LeNet \Te{}s on the corresponding datasets with \textbf{balanced} mini-batches. Trained \Te{}s report test accuracies of $99.44\%$ and $93.38\%$ respectively. 
Note that the test sets comprise equal number of majority and minority class samples from the corresponding original test data. Since there would be more test datapoints for the majority class, we picked samples equal to that in the minority class test set.
Also, in order to ensure maximal diversity within the `majority' test set, equal number of samples from the constituent MNIST (or FMNIST) labels ($3$ to $9$) are considered. 

We then conduct distillation with (i) random (unbalanced), and (ii) balanced arbitrary transfer sets and present the \St's accuracy on the test sets in Table~\ref{tab:unbalanced}. Note that the transfer sets of varying size (from $500$ to $16000$) are composed from the SVHN dataset and the accuracies are reported across $20$ runs. We can clearly observe that the target-balanced arbitrary transfer sets outperform the randomly composed counterparts in vast majority of the cases, thereby validating the generality of the proposed baseline.
\section{Conclusion}
\label{sec:discussion}
Distillation enables a low capacity model (\St{}) to learn a sophisticated mapping which is not possible otherwise (via normal cross entropy training). In order to cope with the constrained operational conditions, recent efforts ~\cite{zskt-neurips-2019,dafl-iccv-2019,zskd-icml-2019,degan-aaai-2020} attempt to distill in a data-free scenario via artificially generated transfer set. Despite using out-of-distribution samples that are visually far away from the actual training data, these methods have reported competitive distillation performance. 
Motivated by these observations, in this work, we explore (i) if a simple baseline can be obtained for data-free KD by leveraging publicly available arbitrary data, and (ii) whether this baseline can be an alternative to the substantially more complex approaches. 
Further, we presented a simple strategy based on intuitive hypothesis to maximize the transfer performance of such sets. Upon extensively experimenting with multiple datasets and model architectures, we bring out the following observations:
\begin{itemize}
    \item Arbitrary (unrelated to the target data) transfer sets can be leveraged to deliver competitive KD performance, when compared with the computationally expensive state-of-the-art data-free distillation methods. 
    Thus, such transfer sets can lead to the design of important baselines for the data-free knowledge distillation task. 
    \item For any arbitrary transfer set, being `target class-balanced' maximizes the transfer performance.
    \item Though class-balancing improves the transfer performance, it depends on the similarity of the transfer set to the original training data. In other words, as the transfer set lies closer to the target data manifold, knowledge transfer improves. (Please refer to the supplementary materials).
\end{itemize}
Hinton \textit{et al.}~\cite{hinton2015distilling} attributed the effectiveness of distillation process to the dark knowledge extracted out of the training data. However, it is very intriguing to understand how even a completely unrelated transfer set with only the distillation objective can help a \St{} to achieve competitive generalization. Also, in the data-free KD setting, it needs to be investigated how the similarity of an arbitrary dataset with the target data distribution can be estimated, especially when multiple such datasets are available for facilitating knowledge distillation. 
We leave these two aspects of the data-free distillation for future research.

{\small
\bibliographystyle{ieee_fullname}
\bibliography{kd}
}


\onecolumn
\begin{center}
    \Large{\textbf{\textit{Supplementary Material for}\\ Effectiveness of Arbitrary Transfer Sets for Data-free Knowledge Distillation} }
\end{center}

\setcounter{section}{0}
\setcounter{table}{0}
\setcounter{figure}{0}
\vspace{8pt}
\hrule
\vspace{18pt}
\section{Importance of Class Balance: Training Data}
\label{subsec:balance-training-data}
We considered AlexNet CNN trained on CIFAR-10 as the \Te{} and AlexNet-Half as the \St{}. We carefully composed multiple transfer sets representing only a subset of classification regions learned by the \Te{} in order to perform the distillation. In other words, we performed distillation using the samples from different number of classes. We varied the number of classes present in the transfer set, fixing its size (total number of samples). 

More specifically, we fixed the size of the transfer set to approximately\footnote{Training data per each class is not exactly equal to $5000$ but very close to it.} $10000$ and varied the class composition from $2$ to $10$. Note that as the number of representing classes increases, number of samples per class decrease to meet the fixed size criterion. Table~\ref{tab:training-data-balance} shows the distillation performance of the transfer sets in terms of the \St{} classification accuracy on the test set that consists of samples from all the $10$ classes. Clearly, the performance increases monotonically with the class balance in the transfer set. In other words, with better representation of the classification regions the transfer set achieves better distillation between the \Te{} and \St{}.
\begin{table}[htp]
\centering
\begin{tabular}{|l|c|c|c|c|c|}
\hline 
\multicolumn{1}{|l|}{\multirow{2}{*}{}} & \multicolumn{5}{c|}{\# classes in the transfer set}                                                                         \\ \cline{2-6} 
\multicolumn{1}{|l|}{}                  & \multicolumn{1}{c|}{2} & \multicolumn{1}{c|}{4} & \multicolumn{1}{c|}{6} & \multicolumn{1}{c|}{8} & \multicolumn{1}{c|}{10} \\ \hline 
Student Acc.       &  19.56        &  34.50        &     46.08    &    58.94     &    71.24      \\ \hline
Feature dist.  &  52.25&49.26 &44.45 &34.72 &8.97 \\ \hline         
\end{tabular}
\caption{Importance of the class balance in a fixed-size transfer set when training data is used for transfer. Test accuracy of the \St{} trained via distillation on samples from different number of CIFAR-10 classes. The table also presents the Hausdorff distance between transfer set and training set computed in the feature space.}
\label{tab:training-data-balance}
\end{table}

We can also verify the effectiveness of a transfer set via measuring its similarity (or distance) to the target data distribution. Note that the data on which the \Te{} model is trained and tested is assumed to be sampled from the target data distribution. Therefore, along with the \St{}'s accuracy we also compute the distance between the transfer set and the training set. Since those are sets of images, we compute the Hausdorff distance~\cite{hausdorff-pami-2015} between the corresponding feature sets. We consider the deepest embedding (before the softmax layer) learned by the \Te{} model as the feature. Bottom row of Table~\ref{tab:training-data-balance} shows the computed distances. Note that the distance monotonically decreases as the balance in the transfer set improves. In other words, as the similarity of the transfer set to the target set improves, the distillation performance improves. 
\vspace{10pt}
\section{Importance of Class Balance: Arbitrary Data}
\label{subsec:balance-arbitrary-data}
In this subsection we demonstrate the importance of class balance in the case of arbitrary transfer set. We consider setup similar to that in section~\ref{subsec:balance-training-data} with AlexNet trained on CIFAR-10 as \Te{} and AlexNet-Half as \St{}. However, here we consider an arbitrary transfer set composed with samples from SVHN and TinyImageNet datasets. Similar to the previous subsection, we fixed the size of the transfer set approximately to $10000$ and analysed the effect of class balance. We varied the class representation from $2$ to $10$ in the transfer set and investigated the distillation performance. Table~\ref{tab:arbitrary-data-balance} shows the \St{}'s classification accuracy on the $10000$ CIFAR-10 test set that has almost equal number of samples from all the $10$ classes. Note that the performance monotonically increases with the class-balance in the arbitrary transfer set.

Similar to section~\ref{subsec:balance-training-data} we also verified the similarity of the transfer sets to the target set via measuring the feature similarity. Bottom row of Table~\ref{tab:arbitrary-data-balance} shows the Hausdorff distance measured between the corresponding feature sets. It is evident that with better class-balance in the transfer set, it is more similar to the target dataset and results in better distillation. Thus, sections~\ref{subsec:balance-training-data} and~\ref{subsec:balance-arbitrary-data} clearly support the proposed hypothesis that class-balance improves the distillation performance.

\begin{table}[htp]
\centering
\begin{tabular}{|l|c|c|l|l|c|}
\hline 
\multicolumn{1}{|l|}{\multirow{2}{*}{}} & \multicolumn{5}{c|}{\# classes in the transfer set}                                                                         \\ \cline{2-6} 
\multicolumn{1}{|l|}{}                  & \multicolumn{1}{c|}{2} & \multicolumn{1}{c|}{4} & \multicolumn{1}{c|}{6} & \multicolumn{1}{c|}{8} & \multicolumn{1}{c|}{10} \\ \hline 
Student Acc. &    23.36       &    26.51      &     32.08    &     39.06    &   44.42       \\ \hline
Feature dist.   &   50.25       &     49.99     &    44.73     &   38.65      &  38.65\\ \hline         
\end{tabular}
\caption{Importance of the class balance in the fixed-size transfer set when arbitrary data (SVHN+Tiny Imagenet) is used. Note that the table also presents the Hausdorff distance between the transfer set and the training set computed in the feature space.}
\label{tab:arbitrary-data-balance}
\end{table}

\pagebreak
\section{Base Transfer Set Matters}
\label{sec:base-transfer-set}
\begin{table*}[htp]
\centering
\begin{tabular}{|l|c|c|}
\hline
Transfer set         & Balance & Distillation Performance \\ \hline
SVHN                 &   \xmark      & 40.46                    \\ \hline
SVHN+Tiny Imagenet   &     \cmark{}    & 64.90                    \\ \hline
Tiny Imagenet        &   \xmark      & 66.94 \\ \hline
Tiny Imagenet + SVHN &   \cmark{}      & 70.58                    \\ \hline
\end{tabular}
\caption{Distillation performance with different base transfer sets on the CIFAR-10 dataset. Note that in order to have a fair comparison, amount of transfer set does not exceed the total count of original training data. The training is done without using any augmentations.}
\label{tab:order-matters}
\end{table*}
Along with class balancing, the choice of base transfer set is also an important factor on which the effectiveness of the KD depends. From Table \ref{tab:order-matters}, it can be observed that there is a significant improvement in the distillation performance when a natural dataset such as TinyImageNet is considered as base transfer set in comparison to SVHN. Moreover, balanced TinyImageNet gives further improvement in the accuracy which is approximately $6\%$ more than using balanced SVHN. Therefore, the order in which the arbitrary datasets are mixed to create a transfer set also matters.
\section{Type of Dataset Used for Balancing also Matters}
\begin{table}[htp]
\centering
\begin{tabular}{|l|c|c|c|c|} \hline
          \pbox[c][1 cm][b]{1.8cm} {Unbalanced \\Dataset} &   \multicolumn{4}{c|}{
          \pbox[c][0.75 cm][c]{7cm} {Dataset which is added to balance class count}
          }  \\ \cline{2-5} 
           &  \pbox[c][1.00 cm][c]{1.75cm}{No dataset \\added }   & \pbox[c][1.00 cm][c]{1.75cm}{Random \\Noise} & \pbox[c][1 cm][c]{1.75cm}{Mid-Air \\(Synthetic)} & \pbox[c][1 cm][c]{2.5 cm} {Tiny Imagenet \\(Natural)}
            \\ \hline
            
Random Noise & 14.38 & 31.03 & 38.03 & 67.29 \\ \hline
Clevr & 36.58 & 41.53 & 47.18 & 60.04 \\ \hline
SVHN & 40.46 & 46.77 & 49.65 & 64.90 \\ \hline

\end{tabular}
\caption{Ablation on different types of dataset when mixed with an unbalanced transfer set to increase the target class balance on CIFAR-10 \Te{}. The values represent the distillation performance.}

\label{type-of-dataset}
\end{table}

From Table \ref{type-of-dataset}, we can observe that an unbalanced transfer set when used directly always gives lower distillation performance in comparison to addition of dataset for increasing the class balance. Please note that we always ensure that count of samples in a balanced transfer set does not exceed the count of samples in unbalanced dataset. Also, the maximum number of samples taken from unbalanced transfer set is limited by the amount of original training data which was used for training the \te{} model in oder to have a fair comparison. This shows that class-balanced transfer sets are more effective than unbalanced transfer set. Even when random noise is added to unbalanced transfer set like random data, synthetic data or natural data improves the distillation accuracy significantly. Moreover, we can notice that the distillation performance also depends on type of dataset which is being added for target class balancing. The distillation performance is best when a natural dataset (Tiny Imagenet) is added to several unbalanced transfer sets like Random Noise, Clevr and SVHN data. 

\pagebreak
\section{What Makes a Better Transfer Set ?}
\label{subsec:transfer-set-selection}
It is clear that not all arbitrary transfer sets are equally effective. Even a pair of target class-balanced transfer sets need not be equally effective. Despite the balance, random noise, synthetic, and natural arbitrary transfer sets result in significantly different transfer performances. This naturally rises the question, \textit{In the absence of the training data, what makes an arbitrary transfer set effective?} By now, intuitively one can expect that ``more the similarity of the transfer set to the training set, better the transfer". Although it is not a sufficient condition, it is consistently observed that better similarity results in effective transfer performance. For instance, Table~\ref{tab:transfer-set-seletion} shows this similarity (in terms of distance) against the corresponding transfer performance for random noise, synthetic, and natural data as transfer set. As the distance at which the transfer set lies from the manifold of the training dataset, its effectiveness decreases. Note that it is not possible to find such similarity easily in the absence of training data. However, this simple observation can guide and influence the future data-free knowledge transfer objectives that attempt to create proxy transfer set (e.g.~\cite{zskd-icml-2019,zskt-neurips-2019,dafl-iccv-2019}).

\begin{table}[htp]
    \centering
    \begin{tabular}{|l|c|c|}
        
        \hline
        Transfer Set & Distillation Acc. & Feature Distance     \\ \hline \hline
        Random Noise & 67.40 & 36.40 \\ \hline
        Synthetic data & 76.92 & 29.91 \\ \hline 
        Natural data &  79.19 &  28.53    \\ \hline
        
    \end{tabular}
\caption{Distillation performance using different types of transfer sets for distilling the knowledge from CIFAR-10 \Te{}.} 
\label{tab:transfer-set-seletion}
\end{table}
\section{Class Frequencies in the Transfer Set}
In this section we present the class frequencies in the transfer set before and after the balancing. In other words, we show the number of samples in the transfer set that are classified into each of the classes in the \Te{}'s training data before and after the proposed class-balancing (Algorithm 1 in the main draft). Note that these counts are related to the performances in Table 1 of the main draft without performing augmentation. Tables~\ref{tab:frequencies-mnist}, \ref{tab:frequencies-fmnist}, \ref{tab:frequencies-cifar-10} show the counts for MNIST, FMNIST, and CIFAR-10 datasets with various arbitrary transfer sets. Note that though the achieved balance is not perfect, it is very significant compared to the unbalanced arbitrary transfer set t which results in the distillation performance (Table 1 of the main draft).  

\begin{table*}[htp]
\centering
\begin{tabular}{|c|c|c|c|c|c|c|}
\hline
\multirow{2}{*}{Class Label} & \multicolumn{2}{c|}{\textbf{Random Noise}} & \multicolumn{2}{c|}{\textbf{Synthetic}}                  & \multicolumn{2}{c|}{\textbf{Natural data}}                                    \\ \cline{2-7} 
                             & Unbalanced       & Balanced       &  \pbox[c][1 cm] [c] {10cm}{Unbalanced\\ (Clevr)} &  \pbox[c][1 cm] [c] {10cm}{Balanced\\ (Clevr + Mid-Air)} &  \pbox[c][1 cm] [c] {10cm}{Unbalanced\\ (SVHN)} &  \pbox[c][1 cm] [c] {10cm} {Balanced\\ (SVHN+Tiny ImageNet)} \\ \hline \hline
0                            & 933              & 6000           & 2868                  & 4415                    & 3295              & 6000                          \\ \hline
1                            & 233              & 6000           & 3376                  & 5450                     & 6758              & 6000                          \\ \hline
2                            & 3721             & 6000           & 156                 & 6000                     & 1588              & 6000                          \\ \hline
3                            & 4835             & 6000           & 898                  & 6000                     & 1092              & 6000                          \\ \hline
4                            & 5910             & 6000           & 31100                 & 6000                     & 31368             & 6000                          \\ \hline
5                            & 10435            & 6000           & 2026                 & 6000                     & 1424              & 6000                          \\ \hline
6                            & 419              & 6000           & 2117                  & 4745                     & 2828              & 6000                          \\ \hline
7                            & 2886             & 6000           & 5884                 & 6000                     & 6014              & 6000                          \\ \hline
8                            & 29280            & 6000           & 9443                 & 6000                     & 3849              & 6000                          \\ \hline
9                            & 1348             & 6000           & 2132                  & 6000                     & 1784              & 5773                          \\ \hline
\end{tabular}
\label{tab:frequencies-mnist}
\caption{Class frequencies (number of samples in each class) before and after class-balancing various transfer sets for performing KD on the MNIST.}
\end{table*}
\pagebreak
\begin{table*}[htp]
\centering
\begin{tabular}{|c|c|c|c|c|c|c|}
\hline
\multirow{2}{*}{Class Label} & \multicolumn{2}{c|}{\textbf{Random Noise}} & \multicolumn{2}{c|}{\textbf{Synthetic}}                  & \multicolumn{2}{c|}{\textbf{Natural data}}                                    \\ \cline{2-7} 
                             & Unbalanced       & Balanced       & \pbox[c][1 cm] [c]{10cm}{Unbalanced\\ (Clevr)} & \pbox[c][1 cm] [c]{10cm}{Balanced\\ (Clevr + Mid-Air)} & \pbox[c][1 cm] [c]{10cm}{Unbalanced\\ (SVHN)} & \pbox[c][1 cm] [c]{10cm}{Balanced\\ (SVHN+Tiny ImageNet)} \\ \hline \hline
0                            & 2751              & 6000           & 6959                  & 6000                     & 8015              & 6000                          \\ \hline
1                            & 3501              & 6000           & 988                 & 6000                     & 3544              & 6000                          \\ \hline
2                            & 917             & 6000           & 5698                 & 6000                     & 4625              & 6000                          \\ \hline
3                            & 404             & 6000           & 656                 & 6000                     & 4892              & 6000                          \\ \hline
4                            & 263             & 6000           & 40                & 1000                     & 390             & 2240                         \\ \hline
5                            & 99            & 6000           & 551                 & 6000                    & 1113              & 6000                          \\ \hline
6                            & 22614         & 6000           & 1413                & 6000                     & 13500            & 6000                          \\ \hline
7                            & 0             & 6000           & 0                 & 92                       & 20                & 178                          \\ \hline
8                            & 29451     &    6000           & 43681                 & 6000                     & 22578              & 6000                          \\ \hline
9                            & 0             & 6000           & 14                 & 844                     & 1323              & 2702                          \\ \hline
\end{tabular}
\caption{Class frequencies (number of samples in each class) before and after class-balancing various transfer sets for performing KD on the FMNIST.}
\label{tab:frequencies-fmnist}
\end{table*}

\begin{table*}[htp]
\centering
\begin{tabular}{|c|c|c|c|c|c|c|}
\hline
\multirow{2}{*}{Class Label} & \multicolumn{2}{c|}{\textbf{Random Noise}} & \multicolumn{2}{c|}{\textbf{Synthetic}}                  & \multicolumn{2}{c|}{\textbf{Natural data}}                                     \\ \cline{2-7} 
                             & Unbalanced       & Balanced       & \pbox[c][1 cm] [c]{10cm}{Unbalanced\\ (Clevr)} & \pbox[c][1 cm] [c]{10cm}{Balanced\\ (Clevr + Mid-Air)} & \pbox[c][1 cm] [c]{10cm}{Unbalanced\\ (SVHN)} & \pbox[c][1 cm] [c]{10cm}{Balanced\\ (SVHN+Tiny ImageNet)} \\ \hline \hline
0                            & 0                & 5000           & 11205                 & 5000                     & 1719             & 5000                          \\ \hline
1                            & 0                & 5000           & 461                  & 2490                     & 34                & 2493                          \\ \hline
2                            & 15               & 5000           & 1416                  & 5000                     & 7886              & 5000                          \\ \hline
3                            & 1                & 5000           & 5937                  & 5000                     & 27073             & 5000                          \\ \hline
4                            & 13               & 5000           & 129                 & 5000                     & 1803              & 5000                          \\ \hline
5                            & 0                & 5000           & 22472                  & 5000                     & 7761              & 5000                          \\ \hline
6                            & 49971            & 5000           & 1680                 & 5000                     & 23                & 5000                          \\ \hline
7                            & 0                & 5000           & 2186                 & 4727                     & 2509              & 5000                          \\ \hline
8                            & 0                & 5000           & 138                 & 5000                     & 928              & 4190                          \\ \hline
9                            & 0                & 5000           & 4376                 & 5000                     & 264               & 5000                          \\ \hline

\end{tabular}
\caption{Class frequencies (number of samples in each class) before and after class-balancing various transfer sets for performing KD on the CIFAR-10.}
\label{tab:frequencies-cifar-10}
\end{table*}

\begin{figure*}[htp]
\centering
\centerline{\includegraphics[width=\textwidth]{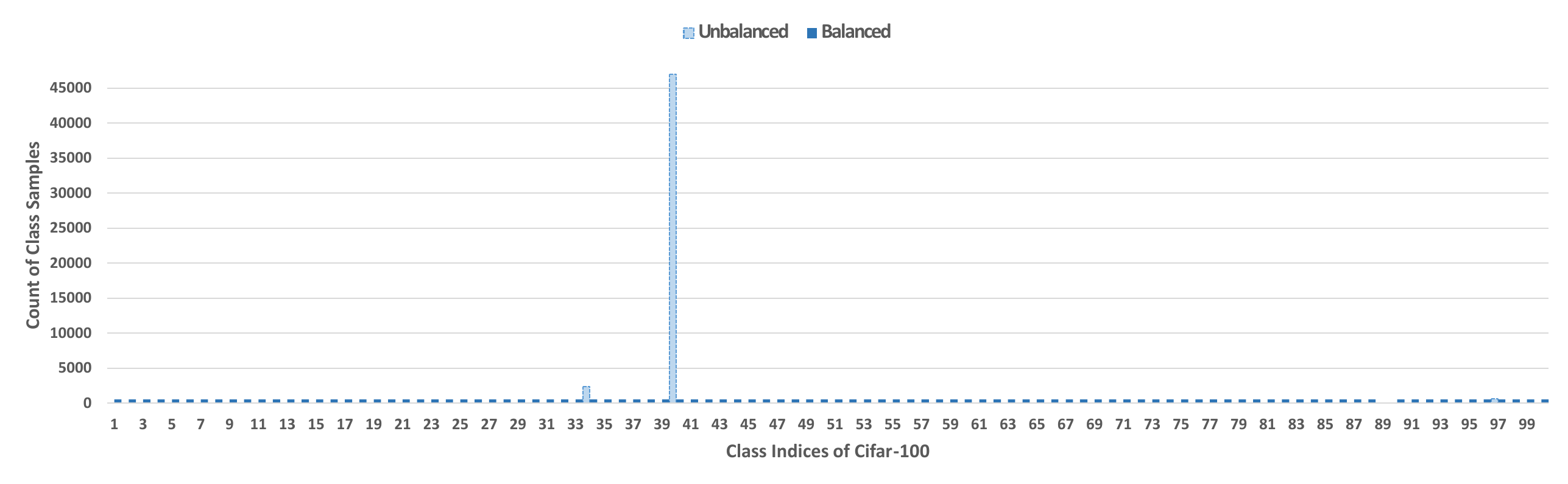}}
\caption{Class frequencies (number of samples in each class) before and after class-balancing the random noise data for performing KD on the CIFAR-100.}
\label{fig:random-cifar-100}
\end{figure*}

\begin{figure*}[htp]
\centering
\centerline{\includegraphics[width=\textwidth]{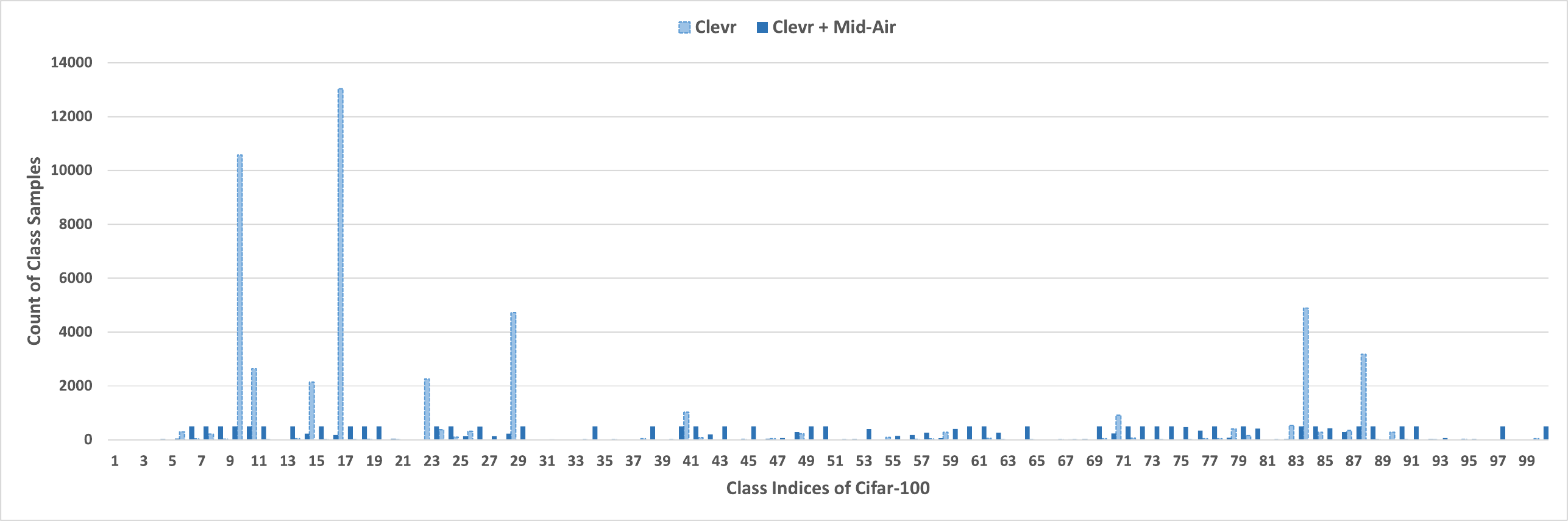}}
\caption{Class frequencies (number of samples in each class) before and after class-balancing the synthetic data for performing KD on the CIFAR-100.}
\label{fig:synthetic-cifar-100}
\end{figure*}

\begin{figure*}[htp]
\centering
\centerline{\includegraphics[width=\textwidth]{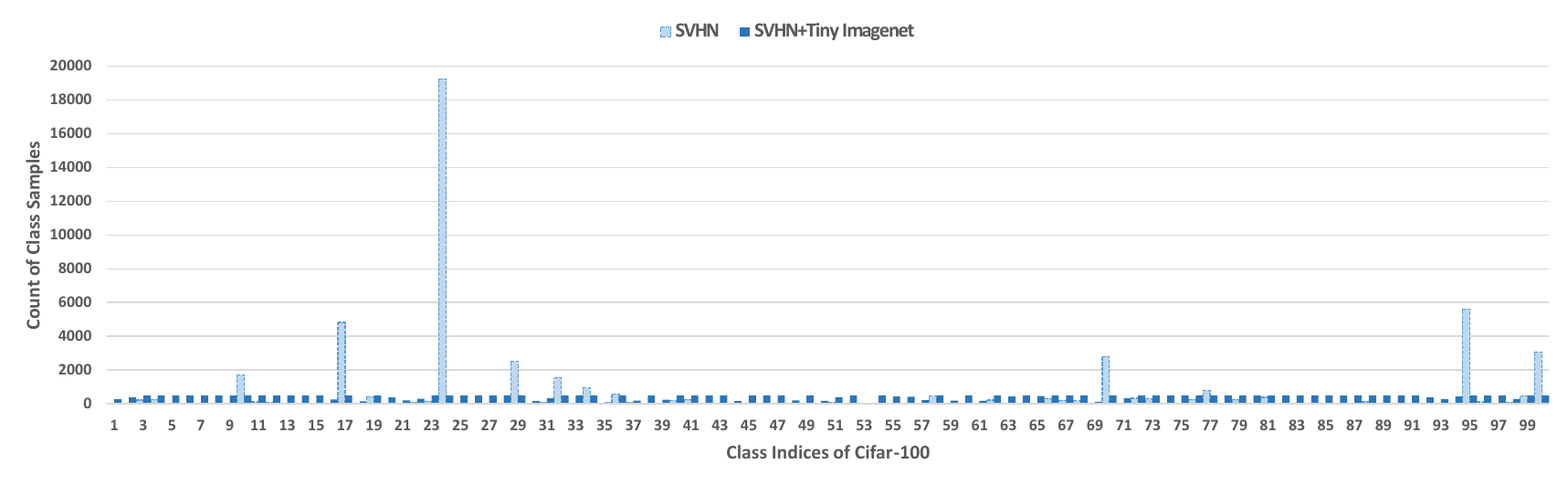}}
\caption{Class frequencies (number of samples in each class) before and after class-balancing the natural data for performing KD on the CIFAR-100.}
\label{fig:natural-cifar-100}
\end{figure*}
\pagebreak
Figures \ref{fig:random-cifar-100}, \ref{fig:synthetic-cifar-100} and \ref{fig:natural-cifar-100} show the class frequencies for CIFAR-100 dataset when the arbitrary transfer sets are Random noise, Synthetic, and Natural datasets respectively. However, note that it is challenging to achieve perfect class balance using any arbitrary transfer set due to the large number of classes present in it. 
Even after mixing `Mid-Air' dataset on the base transfer set `Clevr', it does not achieve the class balance significantly which can also be observed in Figure \ref{fig:synthetic-cifar-100}. In order to retain the same transfer set across several other datasets such as MNIST, FMNIST and CIFAR-10, we did not choose any other synthetic datasets which could have significantly improved the balance of `Clevr' and hence achieved better distillation performance. 

\section{Overall Size of Transfer Sets Used in Experiments}

\begin{table}[htp]
    \centering
    \begin{tabular}{|l|c|c|c|c|c|}
        
        \hline
        Transfer Sets & Balanced & MNIST & FMNIST & CIFAR-10 & CIFAR-100     \\ \hline \hline
        Random Noise & \xmark & 60000 & 60000 & 50000 & 50000 \\ \hline
        Random Noise & \cmark & 60000 & 60000 & 50000 & 50000 \\ \hline
        Clevr & \xmark & 60000 & 60000 & 50000 & 50000 \\ \hline
        Clevr + Mid-Air & \cmark & 56610 & 43936 & 47217 & 26407 \\ \hline
        SVHN & \xmark &   60000 & 60000 & 50000 & 50000 \\ \hline
        SVHN + Tiny Imagenet & \cmark & 59773 & 47120 & 46683 & 43026 \\ \hline

    \end{tabular}
\caption{Total number of samples in transfer sets used for distilling the knowledge from \Te{} model trained on MNIST, FMNIST, CIFAR-10 and CIFAR-100.} 
\label{tab:transfer-set-size}
\end{table}

The amount of original training samples used to train the \Te{} network on MNIST, FMNIST, CIFAR-10 and CIFAR-100 are 60000, 60000, 50000 and 50000 respectively. Due to privacy and safety concerns, we assume the unavailability of these original samples as motivated in several works \cite{zskd-icml-2019, zskt-neurips-2019, dfkd-nips-lld-17}. Thus, in order to train the lightweight models called \St{} network, we leverage on the availability of arbitrary data. This arbitrary data acts as a transfer set for distilling the knowledge of the pretrained \Te{} network. It is evident from the Table \ref{tab:transfer-set-size} that size of arbitrary transfer set does not exceed the amount of original training samples to have a fair comparison. Also, the amount of samples per target class in case of balanced transfer sets, does not exceed the amount of samples per class in original training data which is 6000 for MNIST and FMNIST and 5000 for CIFAR-10 and CIFAR-100 respectively. For the experiments, we limit ourselves to mixture of two arbitrary datasets for obtaining balanced transfer sets. Therefore, we sometimes end up in having non-perfectly balanced transfer sets which have lower transfer set size in comparison to original training samples. Even then, we have shown that these transfer sets achieve better distillation accuracy than the unbalanced transfer sets. 
\vspace{10 pt}
\section{Summary}

Finally, we summarize the major advantages of our proposed approach as follows:
\begin{enumerate}
    \item The proposed method is intuitive in the sense that higher the number of classes and their population, the teacher network is able to transfer more information on to the student network which will reflect in its generalization capabilities.
    \item It achieves competitive distillation performance even with an arbitrary transfer set in the absence of original training data.
    \item It does not require any complicated training procedure, or generative models such as the GANs. The arbitrary transfer sets are used in their original forms (albeit with augmentations applied to them).
    \item Only the distillation loss is used. There is no other additional/auxiliary loss which otherwise needs to be properly weighted with the distillation loss.

\end{enumerate}

Hence the proposed approach can be used as a simple baseline, particularly for Data free Knowledge Distillation research works and can act as an alternative to computationally expensive approaches.

\end{document}